\documentclass[runningheads]{llncs}
\usepackage[T1]{fontenc}
\usepackage{graphicx}
\usepackage{multirow}
\usepackage{floatrow}
\usepackage{marvosym}
\newfloatcommand{capbtabbox}{table}[][\linewidth]

\usepackage{xspace}
\usepackage[namelimits]{amsmath} 
\usepackage{amssymb}             
\usepackage{amsfonts}            
\usepackage{mathrsfs}            
\usepackage{hyperref}
\usepackage{color}

\hypersetup{
    colorlinks=true, 
    linkcolor=blue,  
    filecolor=blue,  
    urlcolor=blue,   
    citecolor=blue,  
}

\newcommand{\our}{{DAPSAM}\xspace}
\usepackage{cleveref}
\usepackage{booktabs}

\begin{document}
\title{Prompting Segment Anything Model with Domain-Adaptive Prototype for Generalizable Medical Image Segmentation}
\titlerunning{Domain-Adaptive Prompt Segment Anything Model}
%
\author{Zhikai Wei \inst{1} \and
Wenhui Dong \inst{1} \and
Peilin Zhou \inst{1} \and
Yuliang Gu \inst{1} \and
Zhou Zhao \inst{2}\and
Yongchao Xu \textsuperscript{1(\Letter)}}

%
\authorrunning{Z. Wei et al.}
%
\institute{$^{1}$ National Engineering Research Center for Multimedia Software, Institute of Artificial Intelligence, School of Computer Science, Medical Artificial Intelligence Research Institute of Renmin Hospital, Wuhan University, Wuhan, China\\
\email{yongchao.xu@whu.edu.cn}\\
$^{2}$ School of Computer Science, Central China Normal University, Wuhan, China\\
}

\maketitle              
\begin{abstract}
Deep learning based methods often suffer from performance degradation caused by domain shift. 
In recent years, many sophisticated network structures have been designed to tackle this problem.  However, the advent of large model trained on massive data, with its exceptional segmentation capability, introduces a new perspective for solving medical segmentation problems. In this paper, we propose a novel \textbf{D}omain-\textbf{A}daptive \textbf{P}rompt framework for fine-tuning the \textbf{S}egment \textbf{A}nything \textbf{M}odel (termed as \textbf{\our}) to address single-source domain generalization (SDG) in segmenting medical images. 
\our not only utilizes a more generalization-friendly adapter to fine-tune the large model, but also introduces a self-learning prototype-based prompt generator to enhance model's generalization ability. Specifically, we first merge the important low-level features into intermediate features before feeding to each adapter, followed by an attention filter to remove redundant information. This yields more robust image embeddings. Then, we propose using a learnable memory bank to construct domain-adaptive prototypes for prompt generation, helping to achieve generalizable medical image segmentation. Extensive experimental results demonstrate that our \our achieves state-of-the-art performance on two SDG medical image segmentation tasks with different modalities. The code is available at~\href{https://github.com/wkklavis/DAPSAM}{https://github.com/wkklavis/DAPSAM}.

\keywords{Single domain generalization \and Medical image segmentation \and Segment Anything Model \and Prompt learning.}
\end{abstract}
\section{Introduction}
The advancement of deep neural networks has led to significant progress in the field of medical image segmentation. Most methods have shown notable performance when the training and testing data share the same distribution. However, distribution shift (also known as domain shift~\cite{domainshift}) leads to a decline in performance, hindering the practical application of deep learning methods in real-world scenarios. In medical image segmentation tasks, this shift occurs more frequently due to discrepancies in imaging distribution caused by non-uniform characteristics of imaging equipment, varying operator skills, and factors such as patient radiation exposure and imaging time. Unlike unsupervised domain adaptation~\cite{uda1} and multi-source domain generalization~\cite{TriD,cddsa}, single domain generalization (SDG) is a more practical but challenging setting, under which only the labeled data from one source domain is used to train the model.

Traditional CNNs mainly focus on style augmentation at the image~\cite{csdg,SLAug,ads} or feature level~\cite{maxstyle,FreeSDG,ccsdg} against domain shifts. CCSDG~\cite{ccsdg} incorporates contrastive feature disentanglement into a segmentation backbone. Recently, Vision Transformers have been shown to be significantly more robust in the out-of-distribution generalization~\cite{rspc,dimix}. In particular, the Segment Anything Model (SAM)~\cite{sam}, trained on more than 1 billion masks, has achieved unprecedented generalization capabilities on a variety of natural images. Some works have shown favorable results when applying SAM to medical image segmentation~\cite{samed,desam,MedSAM,samstudy}. 
DeSAM~\cite{desam} modifies SAM’s decoder to decouple mask generation and prompt embeddings while leveraging pretrained weights, but without fully utilizing the capability and adaptability of the encoder.
These developments showcase the potential of a robust huge segmentation model by leveraging a pre-trained SAM, eliminating the necessity for crafting a complex data augmentation method. 

The accuracy of SAM heavily relies on the design of prompt information, such as dots and boxes.
However, these suitable prompts often require interaction with humans. This type of prompt generation relies on subjective human judgments, often requiring several attempts to find the right prompt.

We introduce a novel prototype-based prompt generation module capable of automatically generating prompts specifically suited for the current image segmentation, which are weakly domain-specific. We aim to generate domain-adaptive prompts by leveraging features learned from the source domain. When confronted with unseen images, we utilize stored feature knowledge to generate instance-level strongly correlated and domain-adaptive prompts that guide the mask decoder in the segmentation process. We implement memory and storage functionality using a parameterized memory bank, taking inspiration from~\cite{MemAE}. Similar to few-shot learning, we aspire to have the module serve as guiding support features when encountering target query features.

To further ensure that the feature information stored in the memory bank is more robust, we redesign a fine-tuning structure to fully harness the model's generalization capability. We use the vanilla adapter structure~\cite{AdaptFormer,medicalsamadapter} to fine-tune the encoder as the basic model. 
Low-level features contain more contour information~\cite{ExplicitVisualPrompt,boundarycue}, which are crucial for medical image segmentation~\cite{unet}. Motivated by this, we propose a new generalized adapter structure, in which low-level information is first mixed with intermediate features.  Then, we further introduce a selective attention mechanism~\cite{cbam,DFF} to suppress information that is detrimental to generalization. After fine-tuning each layer with the generalized adapters in the encoder, we obtain more robust features, further assisting the prompt generation module.

We evaluate the proposed method termed \our on two widely used generalizable medical image segmentation benchmarks. Experiments on different types of datasets show that \our significantly/consistently outperforms previous CNN-based and some other SAM-based methods for single out-of-distribution generalization in medical image segmentation.

Our main contributions are summarized:
    \textbf{1)}We propose a novel domain-adaptive prompt generator using prototype-based memory bank learned from source domain images. This generates domain-weakly-correlated but instance-strongly-correlated prompt, making use of the rich prior knowledge from pre-trained large model for generalization. 
    \textbf{2)}We propose to redesign the adapters in each transformer block by integrating low-level features into intermediate features, followed by a channel attention filter to improve the robustness of image embeddings. 
    \textbf{3)} Extensive experiments show that our \our outperforms previous state-of-the-art methods on two different types of SDG medical image segmentation tasks.

\section{Method}
\label{chapter:method}
The single source domain problem is defined as training on a single source domain $\mathcal{D}^s=\{ x_i^s, y_i^s\} ^{N_s} _ {i=1}$, where $x_i^s$ and $y_i^s$ denote the source image and corresponding label, and then testing model performance on unseen test domains $\mathcal{D}^t=\{ D_1^t, D_2^t,\ldots, D_n^t\}$. 
We use SAM's encoder and decoder as the baseline model. Specifically, we freeze the encoder and adopt two trainable MLP-structured adapters for each layer of the encoder following~\cite{AdaptFormer,medicalsamadapter} for its efficiency and scalability. The decoder is set to fully trainable. Following SAMed~\cite{samed}, we change the original prediction of SAM to semantic segmentation output.



\subsection{Generalized Adapter} 

For an image $I$ of dimension $H \times W$ , we first get the initial image embedding $e_0$ through the frozen Patch Embedding layer of ViT. Then we obtain low-level feature $F_{low}$ from $e_0$ through a simple trainable linear layer.

\begin{figure*}[t]
    \centering
    \includegraphics[width=1.0\textwidth]{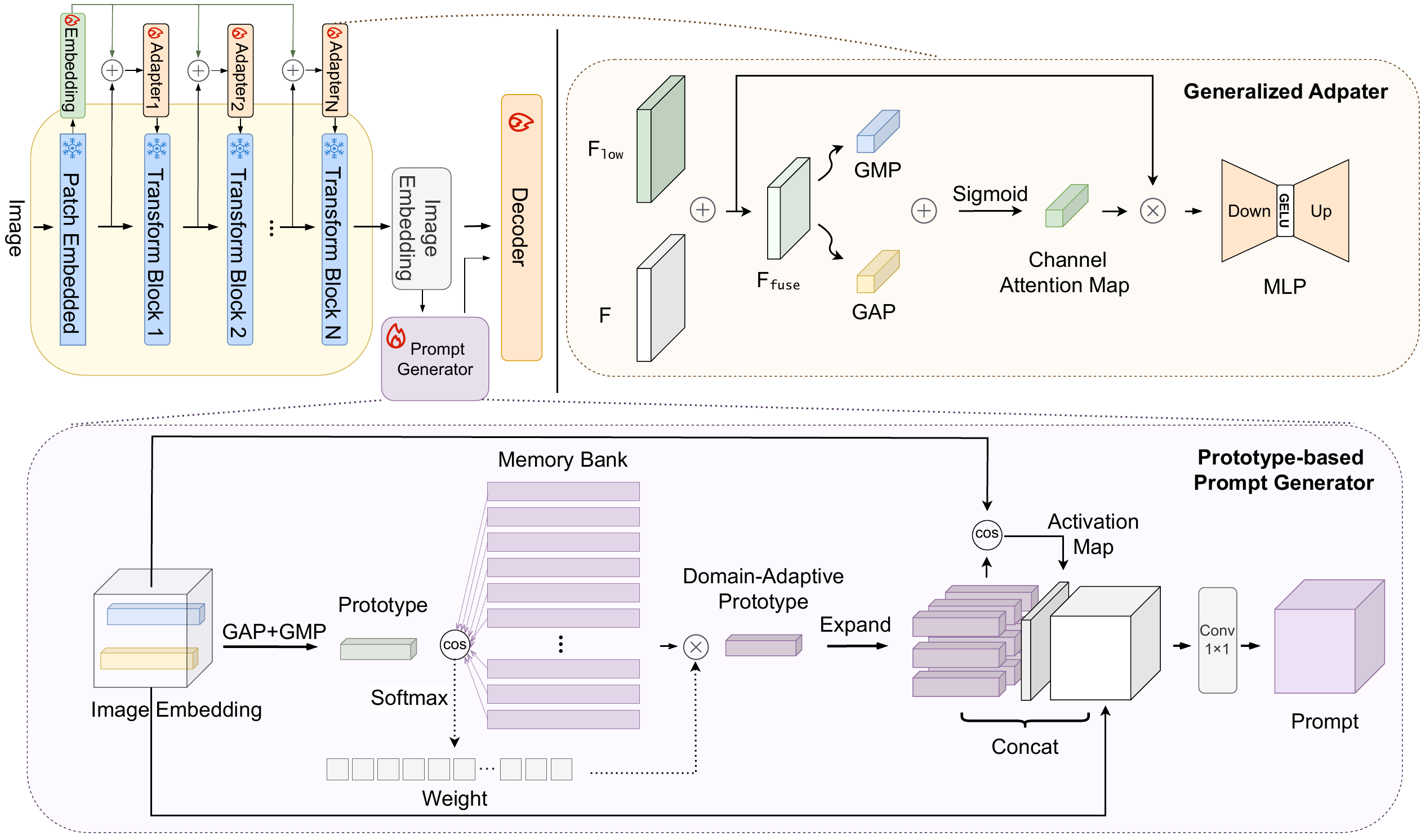}
    \caption{The pipeline of the proposed \our. We design a generalization framework to fine-tune SAM, with generalized adapters (top right) to obtain robust features and a prompt generation module (bottom) to generate instance-related source domain prototypes for target image segmentation.
    }
    \label{fig:Pipeline}
\end{figure*}

In medical images, low-level information such as contours is crucial for the final image segmentation, especially in segmenting organ structures and areas of pathology. We provide low-level feature $F_{low}$ to each intermediate feature $F$ in every adapter. 
We add $F_{low}$ and $F$ to obtain a mixed feature $F_{fuse}$.

Then, we use a selective attention mechanism along the channel dimension to filter out information not conducive to generalization~\cite{DFF}, obtaining more robust features. Channel filtering first involves separately applying global average pooling and global max pooling to the fused feature along the spatial dimensions. The results are then added together, and a sigmoid function is followed to generate a mask which is applied to the fused features. The filtering process can be formulated as: 
\begin{equation}
  F_{filtered} = F_{fuse} \otimes \sigma ( \mathrm{GAP}(F_{fuse}) + \mathrm{GMP}(F_{fuse}) ),
  \label{eq:channelfilter}
\end{equation}
where $\sigma$ denotes the sigmoid function and $\otimes$ denotes element-wise multiplication. $\mathrm{GAP}(\cdot)$ and $\mathrm{GMP}(\cdot)$ respectively denote the global average and max pooling operations along the spatial dimension.

Spatial dimension filtering can disrupt the spatial structure of features, which is often crucial for segmentation. Therefore, different from~\cite{cbam,DFF}, we only employ channel-dimension filtering in our adapters. 

Filtered features are then passed through the vanilla adapter structure, which efficiently and effectively performs adaptation across all layers:
\begin{equation}
  F^{'} = F + \mathrm{MLP_{up}}(\mathrm{GELU}(\mathrm{MLP_{down}}(F_{filtered}))),
  \label{eq:adapter}
\end{equation}
where $F$ represents the original intermediate feature and $F^{'}$ represents the features after adapter. $\mathrm{GELU}(\cdot)$ stands for the GELU activation function, and $\mathrm{MLP_{down}}(\cdot)$ and $\mathrm{MLP_{up}}(\cdot)$ denote the linear layers for downward and upward projection, respectively.

\subsection{Prototype-based Prompt Generator}

After fine-tuning the image through the SAM encoder, the resulting image embedding is input into the subsequent mask decoder. Inspired by prompt learning~\cite{vpt,sam}, we propose to utilize a learnable memory bank to store robust information.
Since prototype vector can capture global information from the feature map and form a continuous semantic space~\cite{pl}, we adopt prototype generated from embedding to interact with the memory bank. 

Firstly, we employ global average and max pooling on the embedding to obtain instance-level prototype $p_i$. For each image $x_i$ embedded into $e_i$, $p_i$ is given by:
\begin{equation}
  p_i =  \mathrm{GAP}(e_i) + \mathrm{GMP}(e_i),
  \label{eq:prototype}
\end{equation}

The memory bank is designed as a parameterized matrix ${M} \in \mathbb{R}^{N \times C}$ with random initialization, where $N$ represents the number of prototypes in the memory bank, and $C$ represents their dimension. Given a prototype vector $p_i \in \mathbb{R}^{1 \times C}$  of an image embedding, the memory bank module utilizes stored knowledge to generate a domain-adaptive and robust prototype $\widehat{p_i}$:
\begin{equation}
  \widehat{p_i} =  p_i \cdot {M} = \sum_{j=1}^{N} \mathit{w_{i, j}} {m_j},
  \label{eq:mbgenerate}
\end{equation}
where ${m_j}$ represents the j-th prototype in the memory bank, and $\mathit{w_{i, j}}$ represents the similarity weight between the prototype $p_i$ of the image and $\mathrm{m_j}$.

We compute each weight $w_{i, j}$ via a softmax operation:
\begin{equation}
  \mathit{w_{i, j}} = \frac{\mathrm{exp}(\mathrm{Sim}(p_i,{m_j}))} {\sum_{j=1}^{N} \mathrm{exp}(\mathrm{Sim}(p_i,{m_j}))}, 
  \quad  \mathrm{Sim}(p_i,{m_j}) = \frac{p_i \cdot {m_j}^{\mathsf{T}}}{\Vert p_i \Vert \Vert {m_j} \Vert},
  \label{eq:mbweight}
\end{equation}
where $\mathrm{Sim}(\cdot,\cdot)$ denotes the cosine similarity operation.

For each image, $\widehat{p_i}$ is the adjusted and more robust prototype feature after being updated through the memory bank. To better guide the embedding, we first compute the cosine similarity between $\widehat{p_i}$ and $e_i$ to generate an activation map ${A}_i$ as the guidance information:
\begin{equation}
  {A}_i = \mathrm{Sim}(\zeta_{h \times w}(\widehat{p_i}), e_i) ,
  \label{eq:activationmap}
\end{equation}
where $\zeta_{h \times w}(\cdot)$ expands the given vector to the same spatial size $h\times w$ as $e_i$.

Then, we concatenate $\widehat{p_i}$, $e_i$, and the map ${A_i}$ and input them into a $1\times1$ convolution to generate a specific prompt for image embedding:
\begin{equation}
  {Prompt_i} = \mathrm{Conv_{1\times 1}}([\widehat{p_i}, {A_i}, e_i ]),
  \label{eq:concattoprompt}
\end{equation}
where $\mathrm{Conv_{1\times 1}}(\cdot)$ refers to a convolution layer with a kernel size of 1, which is used to perform dimension reduction.

Overall, we introduce a novel module to store learned information, compute instance-level difference and generate domain-adaptive prompt. We project all target domain knowledge into the latent space and use source domain knowledge of the memory bank to represent them, which helps to align the source and target domain.
This helps to improve the model's generalization ability.

\subsection{Training objective}

Following SAMed~\cite{samed} and TriD~\cite{TriD}, we combine cross entropy loss and dice loss to supervise the entire training process on the source domain:
\begin{equation}
  \mathcal{L} = (1 - \lambda )\mathcal{L}_{CE} + \lambda \mathcal{L}_{Dice},
  \label{eq:lossfunc}
\end{equation}
where $\lambda$ denotes the weight to balance these two loss terms.

\section{Experiments}
\subsection{Experimental Settings}

\begin{table*}[t] \scriptsize
  \centering
  \caption{ Quantitative comparison of our \our and some state-of-the-art single-source domain generalization methods on prostate dataset. The best and second-best are \textbf{bolded} and \underline{underlined}, respectively. Each column represents leave-one-out results for the model trained on the corresponding domain while testing on the other domains. }
  \scalebox{1}{\begin{tabular*}{\textwidth}{@{\extracolsep{\fill}}l|p{1cm}|cccccc|c}
    \toprule
    Method &  Model & A & B  & C  & D  & E  & F & Average \\
    \midrule
    Upper bound~\cite{prostateupperbound} & U-Net & 85.38 & 83.68 & 82.15 & 85.21 & 87.04 & 84.29 & 84.63 \\
    \midrule
    AdvBias~\cite{advbias} & \multirow{6}{*}{U-Net} & 77.45 & 62.12 & 51.09 & 70.20 & 51.12 &50.69 & 60.45 \\
    RandConv~\cite{randconv} & & 75.52 & 57.23 & 44.21 & 61.27 & 49.98 & 54.21 & 57.07\\
    MixStyle~\cite{mixstyle}& & 73.04 & 59.29 & 43.00 & 62.17 & 53.12 & 50.03 & 56.78 \\
    MaxStyle~\cite{maxstyle}& & 81.25 & 70.27 & 62.09 & 58.18 & 70.04 & 67.77 & 68.27 \\
    CSDG~\cite{csdg}  & & 80.72 & 68.00 & 59.78 & 72.40 & 68.67 & 70.78 & 70.06 \\
    CCSDG~\cite{ccsdg}  & & 80.62 & 69.52 & 65.18 & 67.89 & 58.99 & 63.27 & 67.58 \\
    \midrule
    DeSAM~\cite{desam}[whole] & \multirow{3}{*}{ViT}  & 82.30 & 78.06 & 66.65 & 82.87 & 77.58 & 79.05 & 77.75 \\
    DeSAM~\cite{desam}[grid]  & & 82.80 & 80.61 & 64.77 & 83.41 & 80.36 & \textbf{82.17} & \underline{79.02} \\
    SAMed~\cite{samed} & & 80.42 & \textbf{81.44} & \underline{66.75} & 82.09 & 80.19 & 80.17 & 78.51 \\
    \midrule
    Baseline & \multirow{2}{*}{ViT} & \underline{84.42} & 79.79 & 64.83 & \underline{83.49} & \underline{80.50} & 80.18 & 78.87 \\
    \our(Ours) &   & \textbf{86.34} & \underline{81.05} & \textbf{70.81} & \textbf{85.28} & \textbf{82.91} & \underline{81.48} & \textbf{81.31} \\
    \bottomrule
  \end{tabular*}
  }
  \label{tab:prostate}
\end{table*}

\noindent\textbf{The prostate dataset.} 
The prostate dataset~\cite{prostate} comprises 116 MRI cases from six different domains, namely A: RUNMC, B: BMC, C: I2CVB, D: UCL, E: BIDMC, and F: HK. These cases were collected from three distinct public datasets used for the purpose of prostate segmentation. The slices are resized to a uniform 384$\times$384 resolution with consistent voxel spacing. We employ the Dice Similarity Coefficient (DSC) for the evaluation.

\noindent\textbf{The RIGA+ dataset.}
The multi-domain joint OC/OD segmentation dataset RIGA+~\cite{riga1,riga2,riga3} is used in this paper. This dataset encompasses annotated fundus images from five distinct domains: BinRushed, Magrabia, BASE1, BASE2 and BASE3. For our segmentation model, we select BinRushed and Magrabia as the source domains for training and subsequently evaluate the model's performance on the remaining three domains regarded as target domains. The DSC is also employed as the metric to quantify the segmentation quality.

\noindent\textbf{Implementation Details: }
The rank of the adapter is set to 4 for both efficiency and performance optimization. All training is conducted using the `ViT-B' version of SAM. The initial learning rate is set to $5e^{-4}$, and the weight decay for the AdamW optimizer is set to 0.1. We also adopt the warm-up strategy following SAMed~\cite{samed}, with warm-up periods set to 250 and 25 for the prostate and RIGA+ datasets respectively, due to different data-training settings. We apply early stop at 160 epochs, with a maximum of 200 epochs. The hyperparameter $\lambda$ in Eq.~\eqref{eq:lossfunc} is set to 0.8. The baseline of our method is described at the begining of Section~\ref{chapter:method}.

\subsection{Comparison with SOTA Methods} 

\begin{table*}[t]    \scriptsize
  \centering
  \caption{ Quantitative comparison of our \our and some state-of-the-art domain generalization methods on RIGA+ dataset. The best and second-best are \textbf{bolded} and \underline{underlined}, respectively. In the upper part and lower part, BinRushed [Rows 3-13] and Magrabia [Rows 14-24] are used as the corresponding source domain, respectively. We run the proposed \our three times and report the mean and standard deviations.}
  \begin{tabular*}{\textwidth}{l@{}@{\extracolsep{\fill}}cccccccc@{}@{\extracolsep{\fill}}}

    \toprule
    \centering

    \multirow{2}{*}{Method} & \multicolumn{2}{c}{BASE1} & \multicolumn{2}{c}{BASE2} & \multicolumn{2}{c}{BASE3} & \multicolumn{2}{c}{Average} \\
    \cmidrule{2-3} \cmidrule{4-5} \cmidrule{6-7} \cmidrule{8-9}
      & $D_{OD}$ & $D_{OC}$ & $D_{OD}$ & $D_{OC}$ & $D_{OD}$ & $D_{OC}$ & $D_{OD}$ & $D_{OC}$ \\
    \midrule
     CSDG~\cite{csdg} & 93.56\tiny$\pm$0.13 & 81.00\tiny$\pm$1.01 & 94.38\tiny$\pm$0.23 & 83.79\tiny$\pm$0.58 & 93.87\tiny$\pm$0.03 & 83.75\tiny$\pm$0.89  & 93.93 & 82.85 \\
     ADS~\cite{ads} & 94.07\tiny$\pm$0.29 & 79.60\tiny$\pm$5.06 & 94.29\tiny$\pm$0.38 & 81.17\tiny$\pm$3.72 & 93.64\tiny$\pm$0.28 & 81.08\tiny$\pm$4.97  & 94.00 & 80.62 \\   
     MaxStyle~\cite{maxstyle} & 94.28\tiny$\pm$0.14 & 82.61\tiny$\pm$0.67 & 86.65\tiny$\pm$0.76 & 74.71\tiny$\pm$2.07 & 92.36\tiny$\pm$0.39 & 82.33\tiny$\pm$1.24  & 91.09 & 79.88 \\    
     SLAug~\cite{SLAug} & 95.28\tiny$\pm$0.12 & 83.31\tiny$\pm$1.10 & 95.49\tiny$\pm$0.16 & 81.36\tiny$\pm$2.51 & 95.57\tiny$\pm$0.06 & 84.38\tiny$\pm$1.39  & 95.45 & 83.02 \\    
     D-Norm~\cite{dualnorm} & 94.57\tiny$\pm$0.10 & 81.81\tiny$\pm$0.76 & 93.67\tiny$\pm$0.11 & 79.16\tiny$\pm$1.80 & 94.82\tiny$\pm$0.28 & 83.67\tiny$\pm$0.60  & 94.35 & 81.55 \\
     CCSDG~\cite{ccsdg} & 95.73\tiny$\pm$0.08 & 86.13\tiny$\pm$0.07 & 95.73\tiny$\pm$0.09 & \underline{86.28\tiny$\pm$0.58} & 95.45\tiny$\pm$0.04 & \underline{86.77\tiny$\pm$0.19}  & 95.64 & \underline{86.57} \\
    \midrule
     DeSAM~\cite{desam}[w] & 89.33\tiny$\pm$2.53 & 79.68\tiny$\pm$2.42 & 93.44\tiny$\pm$0.89 & 82.97\tiny$\pm$0.01 & 91.51\tiny$\pm$1.79 & 82.70\tiny$\pm$1.34  & 91.42 & 81.78 \\
     DeSAM~\cite{desam}[g] & 91.79\tiny$\pm$1.62 & 80.87\tiny$\pm$0.11 & 92.57\tiny$\pm$2.04 & 82.95\tiny$\pm$1.62 & 93.66\tiny$\pm$0.07 & 84.19\tiny$\pm$1.79  & 92.67 & 82.67 \\
     SAMed~\cite{samed} & 95.28\tiny$\pm$0.07 & 84.24\tiny$\pm$0.10 & 94.11\tiny$\pm$0.10 & 80.21\tiny$\pm$0.64 & 94.84\tiny$\pm$0.08 &82.60\tiny$\pm$0.32  & 94.74 & 82.35 \\
    \midrule
     Baseline & \underline{95.86\tiny$\pm$0.18} & \underline{86.30\tiny$\pm$0.53} & \underline{95.96\tiny$\pm$0.26} & 80.90\tiny$\pm$0.30 & \underline{96.32\tiny$\pm$0.23} & 86.33\tiny$\pm$0.34  & \underline{96.05} & 84.51 \\
      \our & \textbf{96.34\tiny$\pm$0.17} & \textbf{88.24\tiny$\pm$0.16} & \textbf{96.10\tiny$\pm$0.10} & \textbf{86.31\tiny$\pm$0.13} & \textbf{96.34\tiny$\pm$0.14} & \textbf{88.77\tiny$\pm$0.21}  & \textbf{96.26} & \textbf{87.87} \\
    \hline \hline \addlinespace
     CSDG~\cite{csdg} & 89.67\tiny$\pm$0.76 & 75.39\tiny$\pm$3.22 & 87.97\tiny$\pm$1.04 & 76.44\tiny$\pm$3.48 & 89.91\tiny$\pm$0.64 & 81.35\tiny$\pm$2.81  & 89.18 & 77.73 \\
     ADS~\cite{ads} & 90.75\tiny$\pm$2.42 & 77.78\tiny$\pm$4.23 & 90.37\tiny$\pm$2.07 & 79.60\tiny$\pm$3.34 & 90.34\tiny$\pm$2.93 & 79.99\tiny$\pm$4.02  & 90.48 & 79.12 \\ 
     MaxStyle~\cite{maxstyle} & 91.63\tiny$\pm$0.12 & 78.74\tiny$\pm$1.95 & 90.61\tiny$\pm$0.45 & 80.12\tiny$\pm$0.90 & 91.22\tiny$\pm$0.07 & 81.90\tiny$\pm$1.14  & 91.15 & 80.25 \\ 
     SLAug~\cite{SLAug} & 93.08\tiny$\pm$0.17 & 80.70\tiny$\pm$0.35 & 92.70\tiny$\pm$0.12 & 80.15\tiny$\pm$0.43 & 92.23\tiny$\pm$0.16 & 80.89\tiny$\pm$0.14  & 92.67 & 80.58 \\    
     D-Norm~\cite{dualnorm} & 92.35\tiny$\pm$0.37 & 79.02\tiny$\pm$0.39 & 91.23\tiny$\pm$0.29 & 80.06\tiny$\pm$0.26 & 92.09\tiny$\pm$0.28 & 79.87\tiny$\pm$0.25 & 91.89 & 79.65 \\
     CCSDG~\cite{ccsdg} & 94.78\tiny$\pm$0.03 & 84.94\tiny$\pm$0.36 & 95.16\tiny$\pm$0.09 & 85.68\tiny$\pm$0.28 & 95.00\tiny$\pm$0.09 & 85.98\tiny$\pm$0.29  & 94.98 & 85.53 \\
    \midrule
     DeSAM~\cite{desam}[w] & 82.45\tiny$\pm$2.61 & 69.66\tiny$\pm$2.94 & 84.97\tiny$\pm$0.32 & 75.75\tiny$\pm$1.36 & 83.86\tiny$\pm$2.99 & 74.74\tiny$\pm$2.54  & 83.76 & 73.38 \\
     DeSAM~\cite{desam}[g] & 81.39\tiny$\pm$3.29 & 67.88\tiny$\pm$3.02 & 83.95\tiny$\pm$0.93 & 76.33\tiny$\pm$0.11 & 79.99\tiny$\pm$1.65 & 73.05\tiny$\pm$1.45  & 84.50 & 72.42 \\
     SAMed~\cite{samed} & 95.41\tiny$\pm$0.10 & 85.26\tiny$\pm$0.38 & 95.36\tiny$\pm$0.13 & 84.25\tiny$\pm$0.27 & 95.38\tiny$\pm$0.10 &84.76\tiny$\pm$0.28  & 95.38 & 84.76 \\
    \midrule
     Baseline & \underline{95.50\tiny$\pm$0.33} & \underline{86.63\tiny$\pm$0.22} & \underline{95.88\tiny$\pm$0.24} & \underline{88.29\tiny$\pm$0.35} & \textbf{96.37\tiny$\pm$0.23} & \underline{87.61\tiny$\pm$0.30}  & \underline{95.92} & \underline{87.51} \\
      \our & \textbf{96.22\tiny$\pm$0.18} & \textbf{86.74\tiny$\pm$0.36} & \textbf{96.32\tiny$\pm$0.16} & \textbf{89.59\tiny$\pm$0.24} & \underline{96.35\tiny$\pm$0.20} & \textbf{88.12\tiny$\pm$0.22}  & \textbf{96.30} & \textbf{88.15} \\
    \bottomrule
  \end{tabular*}
  \label{tab:riga}
\end{table*}

\noindent \textbf{Results on prostate} are presented in Table~\ref{tab:prostate}. Compared to the traditional CNN-based U-Net structure, the ViT-based methods designed on SAM show superior performance.  
Our method outperforms the best CNN-based method and some recent SAM-based methods on the prostate dataset. Specifically, compared to the baseline, our method achieves a 2.44\% improvement. Moreover, compared to the recently proposed SAM-based SDG medical segmentation method, DeSAM~\cite{desam}, our approach exhibits a notable 2.29\% enhancement. 

\noindent\textbf{Results on RIGA+} are presented in Table~\ref{tab:riga}. Our DAPSAM still achieves superior performance. When using BinRushed as the source domain, DAPSAM surpasses the CNN-based state-of-the-art CCSDG~\cite{ccsdg} by 0.62\% (96.26\% vs. 95.64\%) and 1.30\% (87.87\% vs. 86.57\%). Our method also outperforms other SAM-based methods and baseline. With Magrabia as the source domain, while the baseline shows impressive results, DAPSAM further improves upon this performance. 
These SOTA results further justify the robustness and competitiveness of our DAPSAM.

\noindent More additional experimental results can be found in the \textbf{supplementary}.

\begin{table*}[t]
\begin{floatrow}
\capbtabbox{ \scriptsize
\renewcommand{\arraystretch}{1.14}
  \begin{tabular*}{0.44\textwidth}{@{\extracolsep{\fill}}cccc|c} 
    \toprule
    \centering Baseline & LLFI & Filter & PPG & Average \\
    \midrule
     \checkmark &   &   &   & 78.87  \\
    \checkmark & \checkmark  &   &   & 79.29  \\
        \checkmark &   & \checkmark  &   & 79.52  \\
    \checkmark & \checkmark  & \checkmark  &   & 79.97  \\
    \checkmark &   &   & \checkmark  & 80.31  \\
    \checkmark & \checkmark  & \checkmark  & \checkmark  & 81.31  \\
    \bottomrule
  \end{tabular*}
}{
 \caption{ Ablation study on the effect of different components on prostate. Our adapter component consists of two parts: low-level feature integration (LLFI) and filtering (Filter). PPG: Prototype-based Prompt Generator. }
  \label{tab:ablationsudy}
}
\capbtabbox{ \scriptsize
  \begin{tabular*}{0.46\textwidth}{@{\extracolsep{\fill}}lccc}
    \toprule
    Num & Params(K) & FLOPs(M) &  Averaged \\
    \midrule
     0 &  0  & 0 & 78.87 \\
     64  & 16 &  75.67 & 79.15  \\
    128 & 32 & 75.71 & 79.42  \\
     \textbf{256}  & \textbf{64} & 75.77 & \textbf{80.31}  \\
      512  & 128 & 75.90 & 79.34  \\
     1024  & 256 & 76.16 & 79.13  \\
      2048 & 512 & 76.69 & 78.92  \\
    \bottomrule
  \end{tabular*}
}{
 \caption{ Ablation study of the memory bank size on prostate segmentation. We vary the number of prototypes stored in the memory bank. The first line is the baseline model without domain-adaptive prompt generator.}
  \label{tab:ablationstudyN}
 
}
\end{floatrow}
\end{table*}

\subsection{Ablation Studies} 
We conduct extensive ablation studies of our method on the prostate dataset.

\noindent \textbf{Effect of different components.}
We first assess the impact of the generalized adapter. As demonstrated in the first to fourth rows of Table~\ref{tab:ablationsudy}, when supplementing only low-level features, the model shows a slight improvement. Notably, using filtering mechanisms to remove redundant information leads to further enhancement. These results reveal that our design not only supplements crucial low-level information in medical image segmentation but also enhances the robustness of intermediate features.

We further explore the role of the Prototype-based Prompt Generator module (PPG). The results presented in the fifth row of Table~\ref{tab:ablationsudy} confirm that incorporating PPG yields a 1.44\% boost in the average Dice score relative to the baseline. This improvement distinctly highlights the PPG module's ability to proficiently utilize the knowledge acquired, thereby significantly enhancing the network's generalization capacity and robustness.

\noindent\textbf{Effect of the memory bank size.}
We evaluate the impact of the hyper-parameter $N$ involved in Eq.~\eqref{eq:mbgenerate}. As depicted in Table~\ref{tab:ablationstudyN}, when $N$ is set to a lower value, suboptimal results implies that a smaller memory bank cannot fully learn all the information. Conversely, an excessively high value of $N$ leads to a decline in performance, since a too large memory bank tends to overfit the source domain information. Optimal performance is achieved for $N$ is set to 256.


\section{Conclusion}
In this paper, we first analyze the performance of fine-tuning large model SAM for domain generalization, and find its excellent potential for generalizable medical image segmentation. We then propose a novel prototype-based domain-adaptive prompt generator to mine such potential of SAM in SDG medical image segmentation. We also propose a more generalization-friendly adapter that improves the robustness of image embedding, further boosting the model's generalization ability. 
The proposed method termed \our outperforms some state-of-the-art CNN-based and SAM-based methods on two widely used benchmarks for generalizable medical image segmentation.

\begin{credits}
\subsubsection{\ackname} This work was supported in part by the National Key Research and Development Program of China (2023YFC2705700), NSFC 62222112 and 62176186, the Postdoctoral Fellowship Program of CPSF (GZC20230924), the NSF of Hubei Province of China (2024AFB245), and CAAI Huawei MindSpore Open Fund.

\subsubsection{\discintname}
The authors have no competing interests to declare that are relevant to the content of this article. 
\end{credits}

%
%
%
%

{
    \small
    \bibliographystyle{splncs04}
    \bibliography{Paper-0929} 
}

\end{document}


\section*{Supplementary}


\begin{figure}[h]
    \centering
    \includegraphics[width=0.8\linewidth]{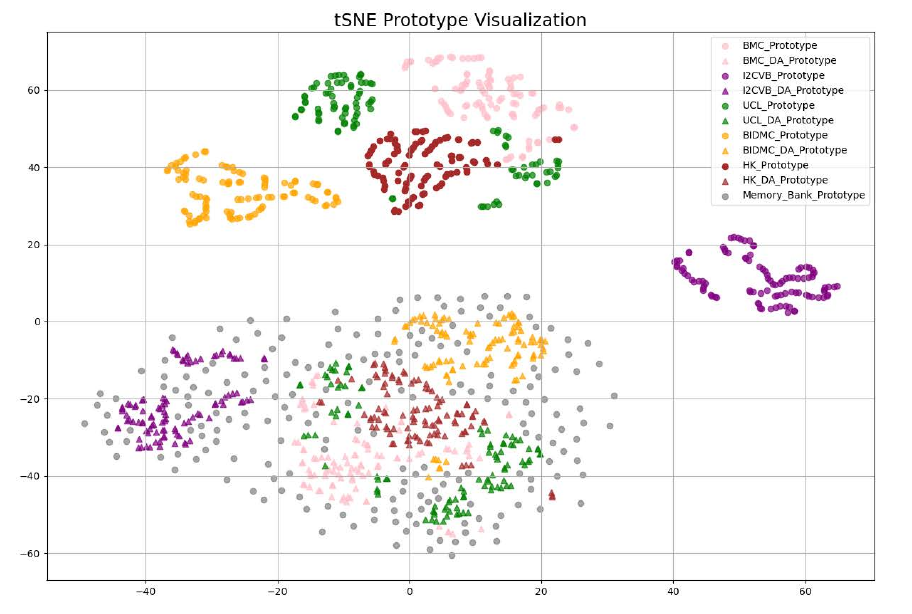}
    \caption{TSNE visualization of prototypes in the memory bank learned from the source domain, original prototypes from test domains, and corresponding domain-adaptive  prototypes. The model is trained on the RUNMC domain of prostate. The other five domains serve as test domains.}
    \label{fig:PrototypeVis}
\end{figure}


\begin{figure*}[t]
\captionsetup[subfloat]{labelsep=none,format=plain,labelformat=empty}
\subfloat[{ Image}]{
\begin{minipage}[b]{0.13\linewidth}
\includegraphics[width=1\linewidth]{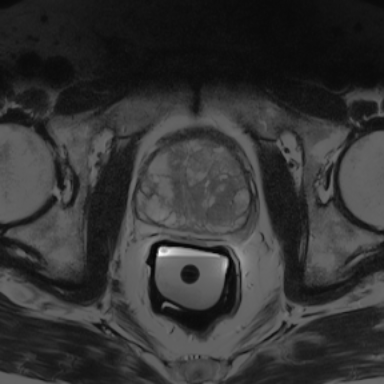}
\includegraphics[width=1\linewidth]{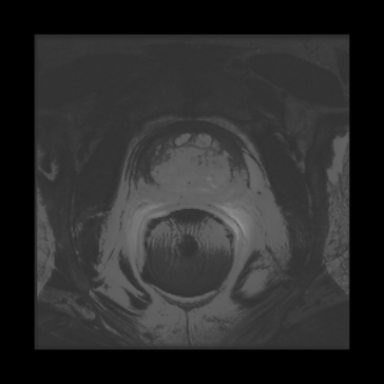}
\includegraphics[width=1\linewidth]{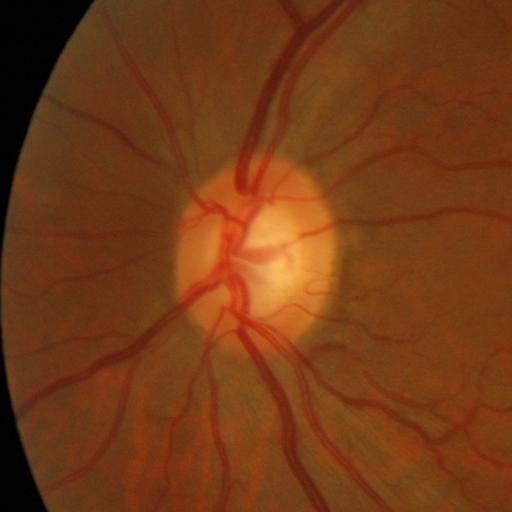}
\includegraphics[width=1\linewidth]{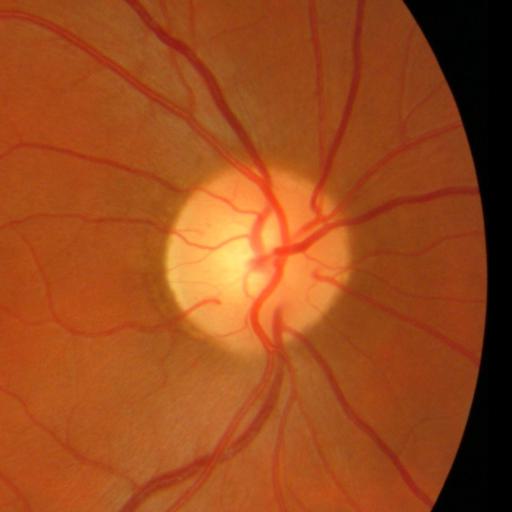}
\end{minipage} 
} 
\subfloat[{ GT}]{
\begin{minipage}[b]{0.13\linewidth}
\includegraphics[width=1\linewidth]{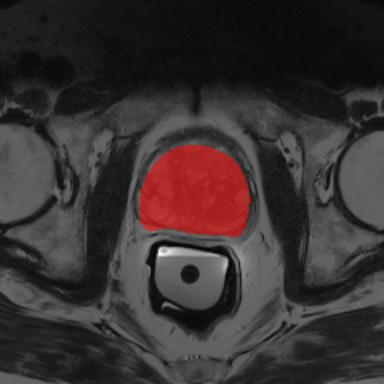}
\includegraphics[width=1\linewidth]{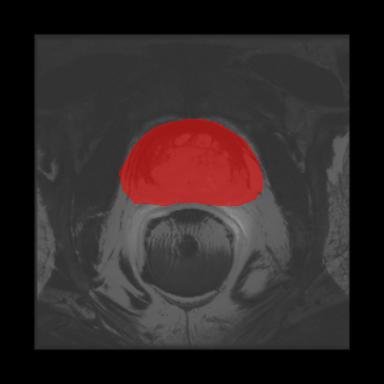}
\includegraphics[width=1\linewidth]{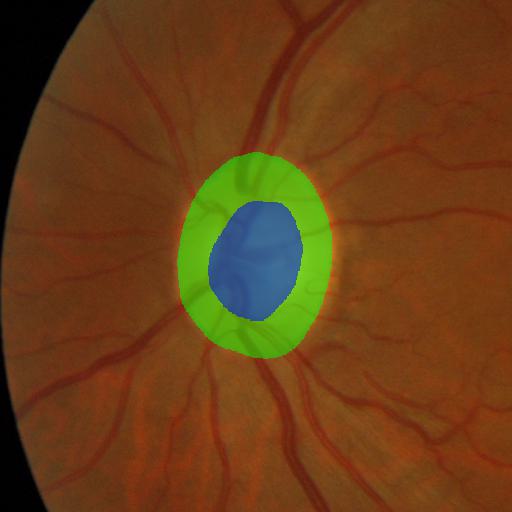}
\includegraphics[width=1\linewidth]{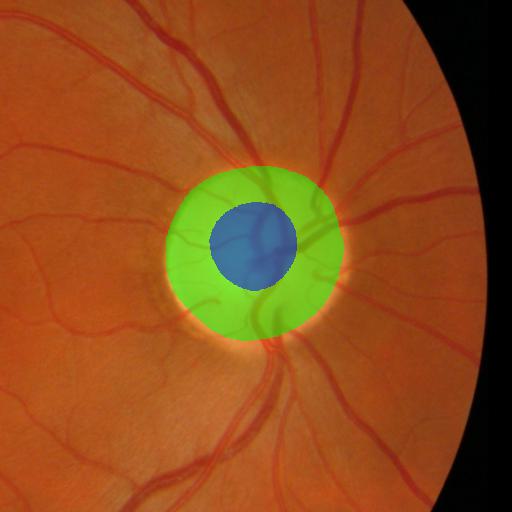}
\end{minipage}  }
\subfloat[{ CCSDG}]{
\begin{minipage}[b]{0.13\linewidth}
\includegraphics[width=1\linewidth]{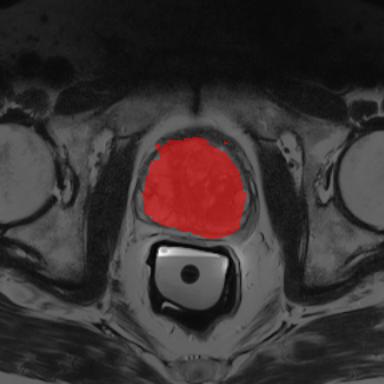}
\includegraphics[width=1\linewidth]{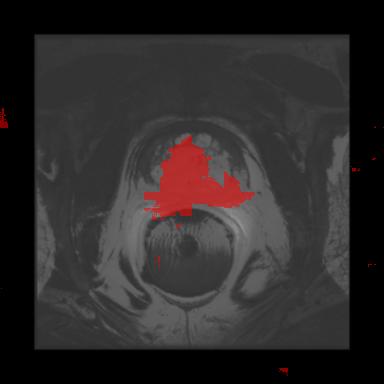}
\includegraphics[width=1\linewidth]{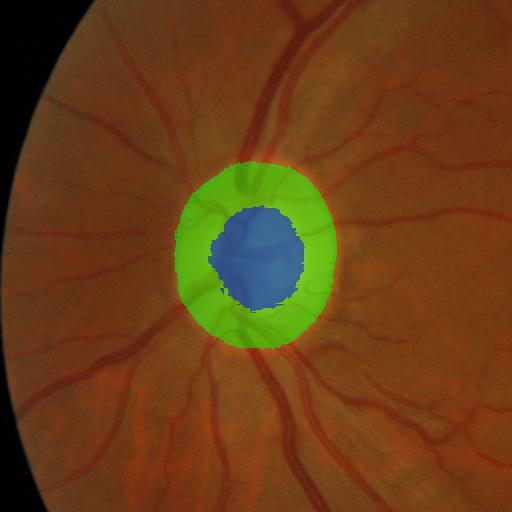}
\includegraphics[width=1\linewidth]{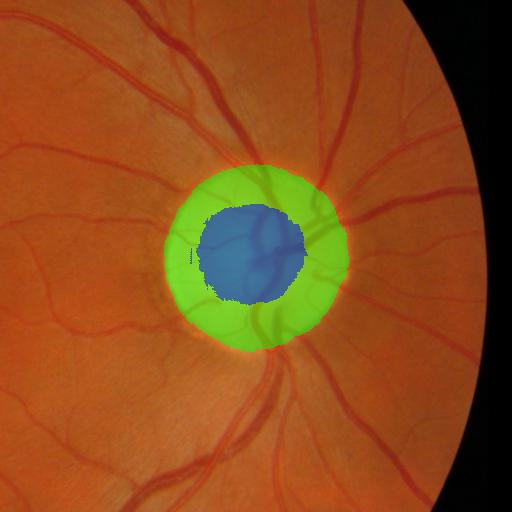}
\end{minipage} }
\subfloat[{ DeSAM[w]}]{
\begin{minipage}[b]{0.13\linewidth}
\includegraphics[width=1\linewidth]{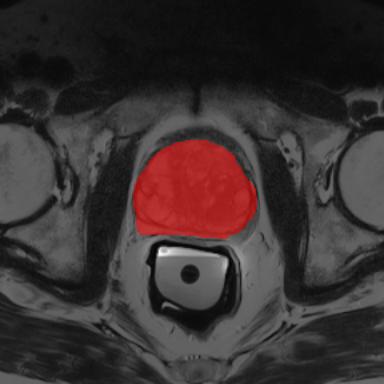}
\includegraphics[width=1\linewidth]{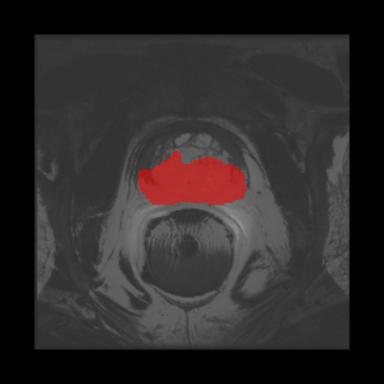}
\includegraphics[width=1\linewidth]{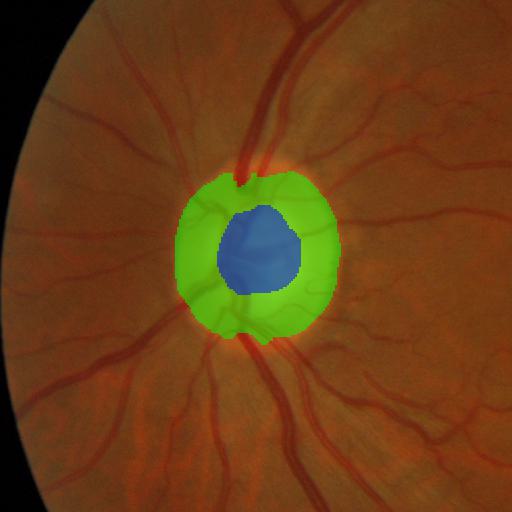}
\includegraphics[width=1\linewidth]{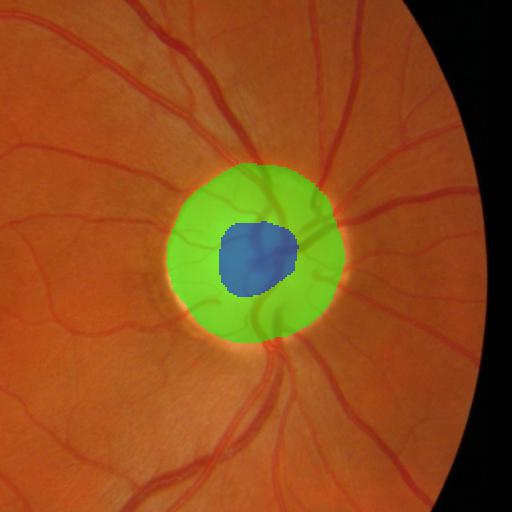}
\end{minipage}  }
\subfloat[{ DeSAM[g]}]{
\begin{minipage}[b]{0.13\linewidth}
\includegraphics[width=1\linewidth]{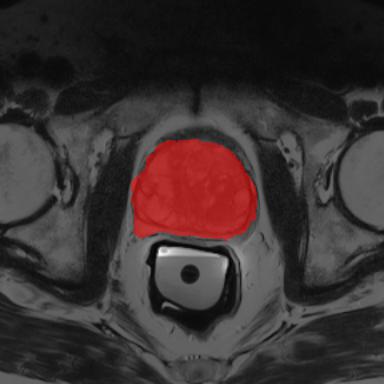}
\includegraphics[width=1\linewidth]{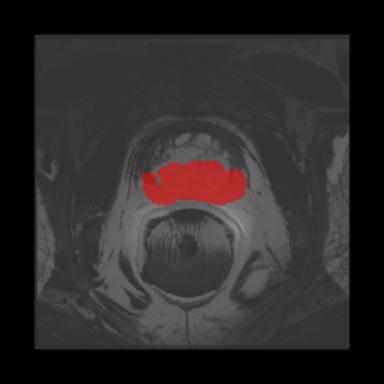}
\includegraphics[width=1\linewidth]{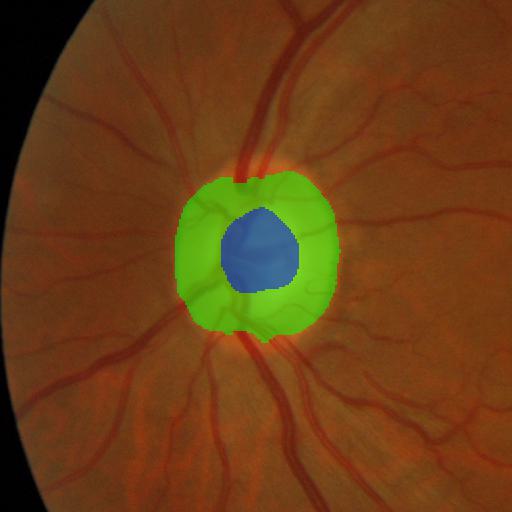}
\includegraphics[width=1\linewidth]{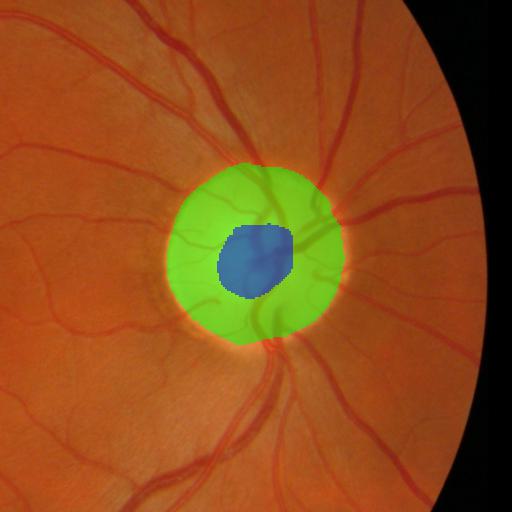}
\end{minipage}}
\subfloat[{ Baseline}]{
\begin{minipage}[b]{0.13\linewidth}
\includegraphics[width=1\linewidth]{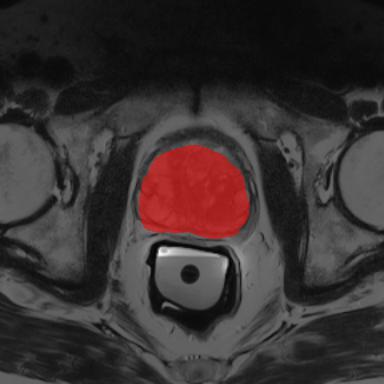}
\includegraphics[width=1\linewidth]{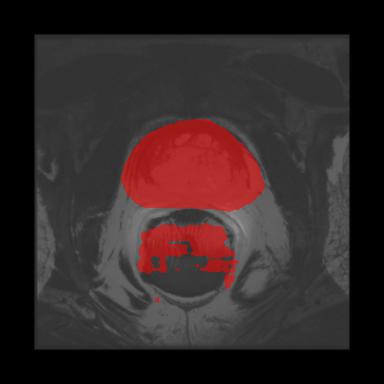}
\includegraphics[width=1\linewidth]{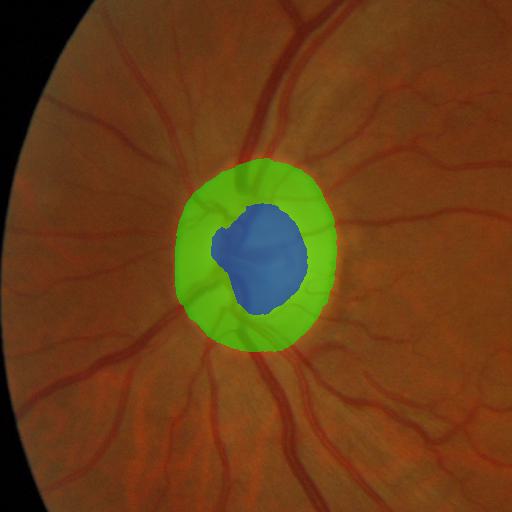}
\includegraphics[width=1\linewidth]{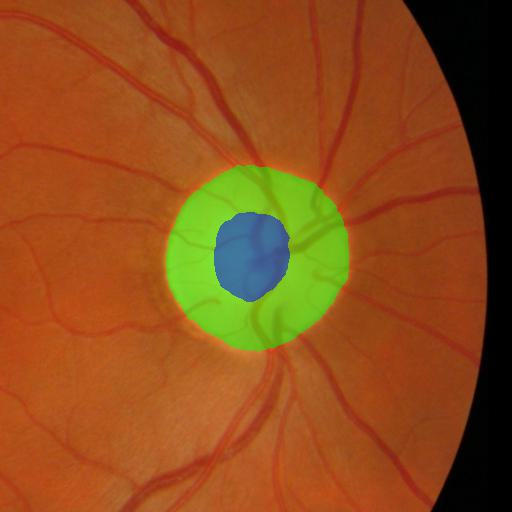}
\end{minipage} }
\subfloat[{Ours}]{
\begin{minipage}[b]{0.13\linewidth}
\includegraphics[width=1\linewidth]{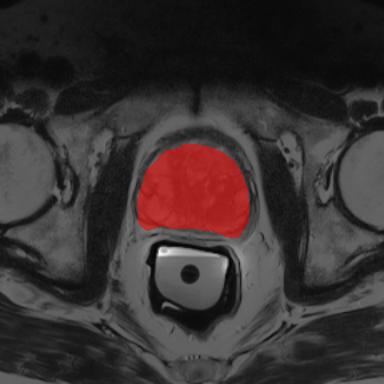}
\includegraphics[width=1\linewidth]{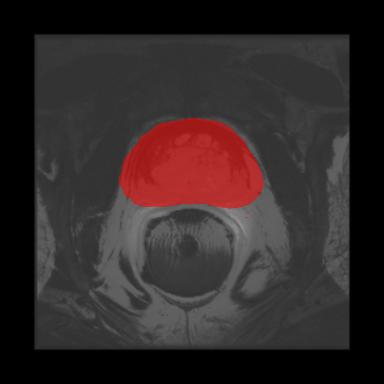}
\includegraphics[width=1\linewidth]{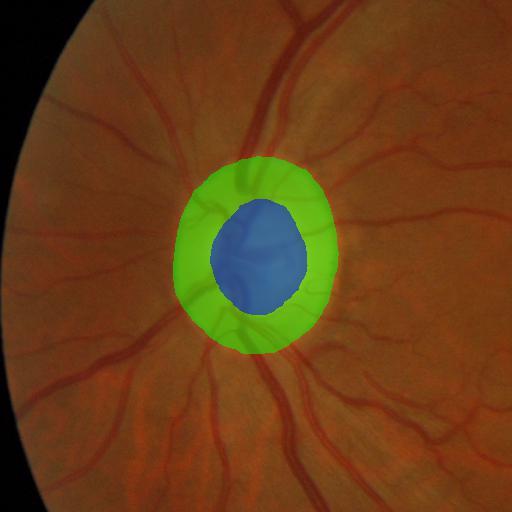}
\includegraphics[width=1\linewidth]{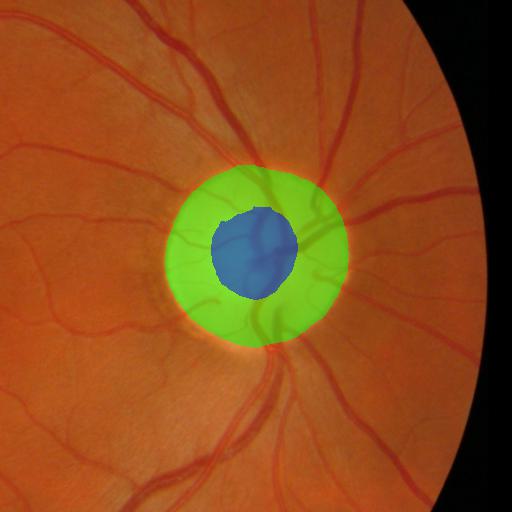}
\end{minipage}}
\caption{ Some qualitative segmentation results on prostate dataset (first two rows) and RIGA+ dataset (bottom two rows). 
}
\label{fig:vis}
\end{figure*}

\begin{table}[h]
  \centering
    \renewcommand{\arraystretch}{1.2}
  \caption{In-Distribution performance of DAPSAM on A domain of prostate dataset.}
 \begin{tabular*}{0.4\textwidth}{@{\extracolsep{\fill}}ccc}
        \hline
        \multirow{2}{*}{Method} & \multicolumn{2}{c}{Intra-A}  \\
        \cline{2-3}
                                & DSC$\uparrow$  & ASD$\downarrow$   \\
        \hline
         Baseline &  93.58  & 0.60  \\ \hline
         DAPSAM   &  \textbf{93.92}  & \textbf{0.49}  \\ \hline
    
      \end{tabular*}
  \label{tab:exp_a_prostate}
\end{table}